\documentclass[10pt,twocolumn,letterpaper]{article}

\usepackage[pagenumbers]{cvpr} 

\usepackage[dvipsnames]{xcolor}

\definecolor{cvprblue}{rgb}{0.21,0.49,0.74}
\usepackage[pagebackref,breaklinks,colorlinks,citecolor=cvprblue]{hyperref}
\usepackage{dsfont}
\usepackage[normalem]{ulem}

\def\paperID{4578} 
\def\confName{CVPR}
\def\confYear{2024}

\title{One2Avatar: Generative Implicit Head Avatar For Few-shot User Adaptation}
\author{
Zhixuan Yu$^{1,2}$\thanks{}
\hspace{10mm}Ziqian Bai$^1$
\hspace{10mm}Abhimitra Meka$^1$
\hspace{10mm}Feitong Tan$^1$
\hspace{10mm}Qiangeng Xu$^1$
\\
Rohit Pandey$^1$
\hspace{10mm}Sean Fanello$^1$
\hspace{10mm}Hyun Soo Park$^2$
\hspace{10mm}Yinda Zhang$^1$
\\
\hspace{-5mm}$^1$Google
\hspace{20mm}$^2$University of Minnesota
}

\begin{document}

\twocolumn[{%
\maketitle
\thispagestyle{empty}
\centering
\includegraphics[width=\textwidth]{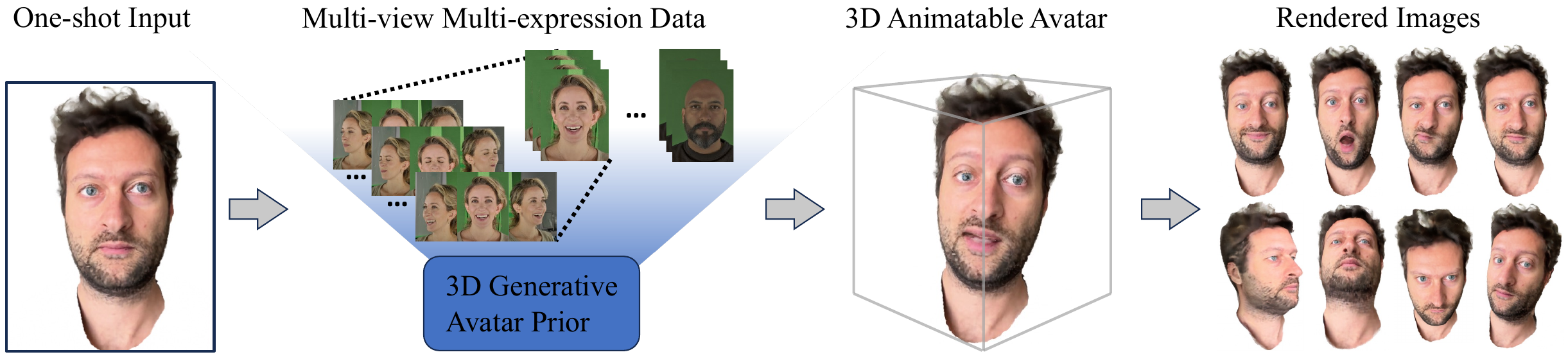}
\captionof{figure}{We present a new approach to generate an animatable photo-realistic avatar from only a few or even one image of the target person. We encode the geometry and appearance by leveraging a neural radiance field that is generated by a 3D generative  model learned from multi-view multi-expression data. With this model, we render the high fidelity head avatar seen from a novel view.} 
\label{fig:teaser}
 \vspace{10mm}
\label{fig:my_label}
}]
{
  \renewcommand{\thefootnote}%
    {\fnsymbol{footnote}}
  \footnotetext[1]{Work was conducted while Zhixuan Yu was an intern at Google.}
}

\begin{abstract}
Traditional methods for constructing high-quality, personalized head avatars from monocular videos demand extensive face captures and training time, 
posing a significant challenge for scalability.
This paper introduces a novel approach to create high quality head avatar utilizing only a single or a few images per user. 
We learn a generative model for 3D animatable photo-realistic head avatar from a multi-view dataset of expressions from 2407 subjects, and leverage it as a prior for creating personalized avatar from few-shot images. 
Different from previous 3D-aware face generative models, our prior is built with a 3DMM-anchored neural radiance field backbone, which we show to be more effective for avatar creation through auto-decoding based on few-shot inputs. 
We also handle unstable 3DMM fitting by jointly optimizing the 3DMM fitting and camera calibration that leads to better few-shot adaptation.
Our method demonstrates compelling results and outperforms existing state-of-the-art methods for few-shot avatar adaptation, paving the way for more efficient and personalized avatar creation. Please see our \href{https://zhixuany.github.io/one2avatar_webpage/}{webpage} for more results.
\end{abstract}

\section{Introduction}
\label{sec:intro}

Creating photo-realistic animatable head avatars is a major enabling factor of authentic social telepresence, which opens up a new opportunity in cross-domains including
communication \cite{Orts-Escolano2016Holoportation,VRChat2023}, embodied AI \cite{XHoloVirtual2023}, and volumetric content \cite{Guo2019The,Meka:2020}.
Among many models, neural radiance fields~\cite{nerf}, in particular, offer a powerful, flexible, and generalizable representation that allows generating 3D photo-realistic head avatars from images. 
This eliminates the need for specialized expertise and hardware to obtain paired 3D annotation \cite{Guo2019The,meta_codec_1,meta_codec_2}. 
However, these methods built upon the neural radiance field models \cite{chen_authentic,meta_codec_1,meta_codec_2} require extensive video sequences of the user capturing diverse facial expressions seen from various viewpoints, followed by a long avatar creation process to achieve a high-fidelity head avatar, which significantly impairs scalability.

One way to mitigate the data requirements is to leverage a prior model learned from large scale data.
Such a data-driven prior has been shown to benefit 3D synthesis for static faces for few-shot inputs \cite{preface, eg3d}.
In the context of learning animatable head avatars, several methods \cite{monoavatar,nerface,rignerf} have explored using parametric models of faces, that are built using large 3D data collections, to better constrain the problem.
However, these methods often struggle to produce high-quality avatars from limited input data, resulting in noticeable animation artifacts and identity shifts. Such prior models have limited expressibility either due to low-dimensionality or due to being trained on collections of single-view images, which limits their ability to learn the subtle variations in facial expressions and appearance from different viewpoints that are crucial for creating realistic and convincing head avatars.
On the other hand, talking heads datasets~\cite{VoxCeleb17,VoxCeleb18,VoxCeleb19} come with a large number of training frames, however these datasets are rarely used for learning avatar prior presumably because they contain limited identities, less expression diversity, and 3D tracking is not easily accessible and stable.

In this paper, we aim to create high quality photo-realistic controllable head avatars from sparse inputs, \eg just a few shots of casually captured images of the subject.
The core of our method is to learn a generative model from various identities and expressions that can serve as a prior for an animatable head avatar, which encodes not only 3D-aware photo-realistic portrait heads but also dynamic deformations correlated with avatar controlling signals, \eg facial expression and head rotations.
Realizing that existing public available data sources are not sufficient, we resort to a customized data capture protocol to collect data that best serves the learning of an avatar prior.
Note that simply adding the missing diversity around camera viewpoint and expression at an equivalent scale of the identity (\eg 70K in FFHQ~\cite{ffhq}) can easily overwhelm the data quantity and lead to an unaffordable capture workload. We found that with a balanced mixture of information across identity, camera, and expression, a reasonable prior model for avatars can be  learnt even from a relatively smaller scale of identities.
In particular, we collect portrait images of $2407$ subjects from $13$ camera viewpoints for up to $13$ types of expressions, resulting in $208K$ images in total.

We use a state-of-the-art avatar representation -- 3DMM-anchored Neural Radiance Field \cite{monoavatar}, and build our generative prior model on it. Our model produces an identity embedding from a learned latent space and an expression embedding from 3DMM parameters, which are mapped to the 3DMM UV space.
These two embeddings are then resampled to the vertices of the 3DMM mesh and  interpreted by MLPs into a continuous neural radiance field.
Our avatar generative prior is learned directly through auto-decoding and supervised using a reconstruction loss.

To create 3D-aware portraits from few-shot images, previous work optimize the latent space of a generative model that captures the user identity, which may involve also fine-tuning of the network parameters~\cite{pti}.
This process is notably unstable and strongly dependent on the specific testing data used.
The additional dependency on facial expression also adds challenges to few-shot adaptation. Poor estimation of camera pose and facial expressions also causes sub-optimal adaptation performance. 
Especially in multi-view setting, inaccurate face fittings lead to misaligned appearance during the optimization, and result in blurry rendering results.
To overcome this, Our method jointly optimizes for all the parameters, including the identity latent code, 3DMM expression parameters, camera pose, and network weights of the neural radiance field decoder.

In summary, our contributions are as follows:

\begin{itemize}
\item We demonstrate that a dynamic photo-realistic head avatar prior model can be effectively learned from a captured dataset of a practical size. 
\item We show that the expression and camera view diversity is essential to such a prior, and should be balanced with identity diversity to form an affordable capture workload.
\item We present a generative head avatar model built with a 3DMM-anchored neural radiance field backbone, which encodes both identity (appearance and geometry) and deformation priors.
\item We propose a joint optimization framework for camera pose and face fitting along with the generative inverse fitting to stabilize the few-shot adaptation.
\end{itemize}

Extensive experiments show that our method enable high quality photo-realistic avatar creation from single or few-shot input images.


\section{Related Work}
\label{sec:related}

\subsection{Video-based 3D Avatar Creation}
Since the introduction of implicit 3D representations like Neural Radience Fields~\cite{nerf}, remarkable results have been shown for creating an animatable 3D head avatar for a target subject from a monocular selfie video sequence capturing various of expressions and head rotations. 
Early approaches derive an avatar directly as a neural radiance field conditioned on the 3DMM parameters~\cite{nerface}, or build an avatar in a canonical space (\eg neutral expression) which is then warped to the target expression based on a 3DMM driven deformation field~\cite{rignerf}.
More recently, another line of work~\cite{monoavatar} attaches local features to 3DMM mesh vertices and establishes a dynamic neutral radiance field that can be deformed with the mesh. This type of geometry-anchored representation explicitly leverages 3D geometry as a prior that offers controllability for human avatars in a natural way. 
However, all of these works require hundreds of subject-specific images for training.
In contrast, we adopt a geometry-anchored representation and augment it with generative capabilities that enable few-shot avatar creation, alleviating the requirement for an extensive video sequence from every target subject, which hinders scalability.

\subsection{Data-driven Prior for 3D Head Synthesis}
A series of works~\cite{diner,keypointnerf,pigan, eg3d, stylenerf, gram,live_3d_portrait,eg3d,preface,chen_authentic} explore 3D-aware portrait generation from few-shot images using generative priors. 
They pre-learn from existing datasets containing a large corpus of subjects, significantly reducing the data required from a specific user. 
One line of work relies on an encoder to extract pixel-aligned features from multi-view images~\cite{keypointnerf,diner}, however, these methods suffer from generalization to novel data due to the missing optimization phase during test time.

On the other hand, 3D-aware generative image synthesis approaches~\cite{pigan, eg3d, stylenerf, gram} learn a prior distribution of human faces from large in-the-wild face image datasets, allowing reconstruction of volumetric radiance fields of novel subjects by performing inversion and fine-tuning of the model~\cite{pti} on one or few images of the target identity.

While offering enhanced quality, they are predominantly trained on large-scale in-the-wild single-view images with a sharp distribution of frontal head pose images, preventing generalization to more extreme camera views.
Preface~\cite{preface} addresses this limitation by learning a prior from a multi-view dataset of diverse neutral faces.
Sharing similar intuitions, we propose to leverage multi-view and multi-expression data to learn a prior for animatable avatars.

Other approaches \cite{chen_authentic} leverage a large scale in-house dataset comprising of millions of images from tens of cameras with registration to build a universal prior. New users are then enrolled via a scanning procedure that requires specialized depth sensors to provide a sufficient level of quality.

\subsection{3D Avatar Generation}
This area of research focuses on developing controllable avatars, where the primary challenge lies in how to effectively model avatar deformations and design the generation backbone.
AniFaceGAN~\cite{anifacegan} extends a 3D-aware image generation method~\cite{gram} for a static head to be animatable by modelling a deformation field mapping query point from observation space to canonical space. OmniAvatar~\cite{OmniAvatar} pre-learns such a mapping from FLAME~\cite{flame} mesh collections while using tri-planes~\cite{eg3d} to represent the 3D feature space of a head in canonical space. Next3D~\cite{next3d} is also built on top of a tri-plane based generation backbone~\cite{eg3d} but utilizes explicit mesh-guided deformation to transfer generated neural textures from UV domain to tri-planes by rasterizing deformed 3DMM mesh. While the focus of these methods is on generating an avatar for novel subjects, they still demonstrate the downstream application for one-shot avatar creation~\cite{next3d} via pivotal tuning inversion (PTI)~\cite{pti}, which aligns with our goal. However, those works are trained only on single-view datasets (\eg FFHQ~\cite{ffhq}) and show limited generalizability towards extreme camera poses and unseen expressions.

\begin{figure*}[t]
\centering
\includegraphics[width=\textwidth]{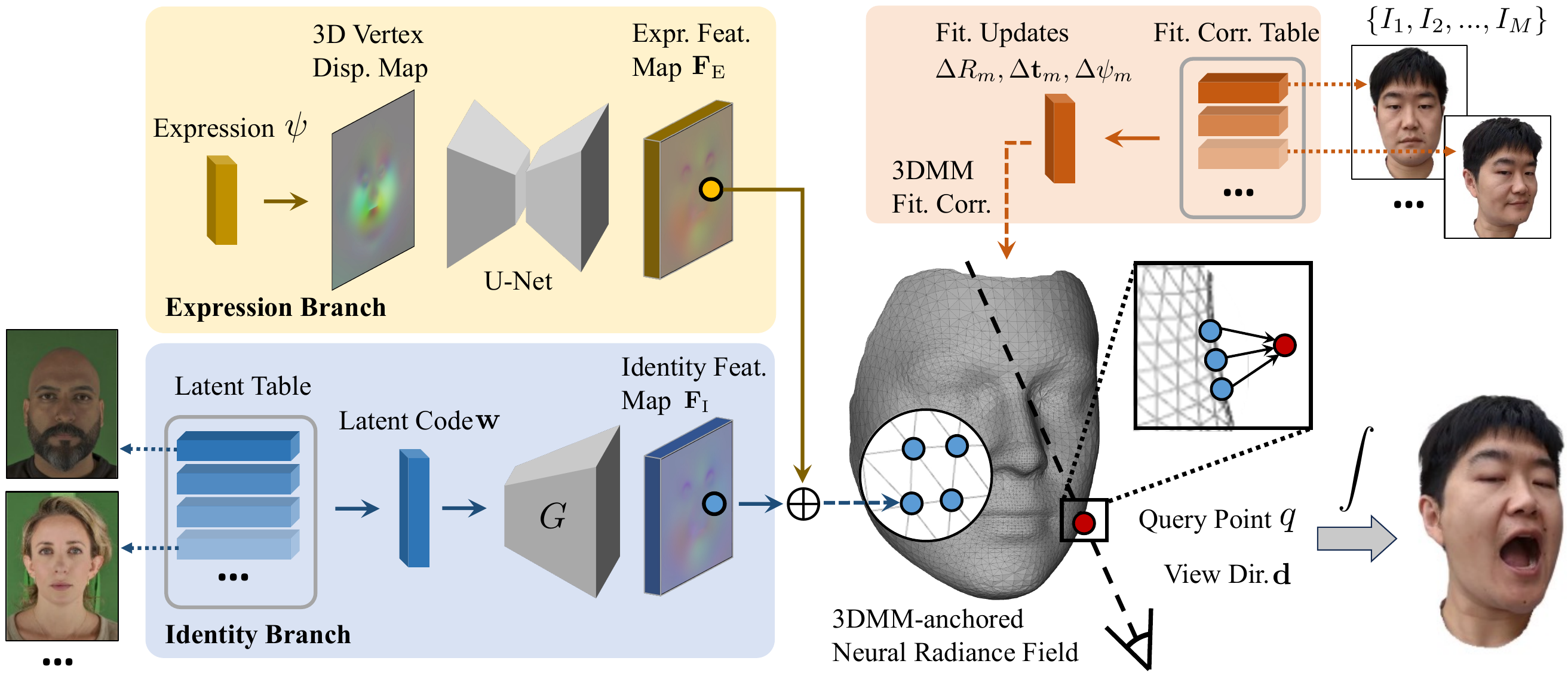}
\caption{We adopt a 3DMM-anchored neural radiance ﬁeld (NeRF) as our avatar representation, where the feature for each query point is aggregated from its k-Nearest-Neighbors in the 3DMM vertices and decoded to color and density via a shallow MLP netowrk. The StyleGAN2~\cite{stylegan2} generator based identity branch encodes personalized characteristics into an identity feature map from a latent code uniquely assigned to a training subject. The expression branch produces an expression feature map from 3DMM expression code via a U-Net. 
The summation of two feature maps are then sampled by 3DMM vertices via texture coordinates to establish the 3DMM-anchored neural radiance field.
For few-shot adaptation, we initialize the target latent code with the mean latent code across training subjects and jointly optimize it with the model weights as well as per-frame 3DMM fitting corrections on given images of the target person.
}
\label{fig:network}
\end{figure*}

\section{Method}
\label{sec:method}

Given a sparse set of $M$ facial images $\{I_1, I_2, ..., I_M\}$ of a person capturing the current appearance ($M$ can be as small as one), our goal is to reconstruct an animatable photo-realistic head avatar that can be rendered from a novel view point.
We learn a generative avatar prior that provides a set of generic features shared across identities and expressions from a multiview dataset of $2407$ identities. Given the prior model, we generalize it to a target identity given $M$ number of images.

\subsection{Multi-view Multi-Expression Face Capture}
We capture high-resolution facial images of a total $2407$ subjects in $13$ pre-defined facial expression from $13$ sparse camera viewpoints.
For each subject at each expression, we run a 3DMM fitting algorithm based on facial landmarks and reconstruct 3D geometry from multi-view images~\cite{3dmm}.
Our dataset includes a limited number of distinctive subjects compared to the existing data, e.g., 70K in FFHQ~\cite{ffhq}. Nonetheless, it includes a wide range of facial expressions, which plays a key role in learning a generative prior model. Please refer to supplementary materials for more details of the capturing setup, including camera placement, pre-defined expressions, and other statistics.

\subsection{Generative Avatar Prior}
Our generative avatar prior generates an avatar represented by a neural radiance field.
Given an identity code $\mathbf{w}$ and an expression code $\mathbf{\psi}$, our model $f$ generates the local color $\mathbf{c}$ and density $\sigma$ for a 3D query point $q$ viewed from direction $\mathbf{d}$:
\begin{equation}
    \sigma(q), \mathbf{c}(q) = f(\mathbf{w}, \mathbf{\psi}, q, \mathbf{d}; \theta),
\end{equation}
where $\theta$ are the model weights.
The color image is then generated by applying a volumetric rendering formulation \cite{nerf} to get per-pixel color:

\begin{equation}
\mathbf{\hat{c}} = \int_{t_n}^{t_f} T(t) \sigma_q(\mathbf{r}(t)) \mathbf{c}_q(\mathbf{r}(t), \mathbf{d})~dt,
\end{equation}
where $T(t) = \exp \left( -\int_{t_n}^{t} \sigma_q(r(s)) \, ds \right)$.
Following prior arts, we adopt the 3DMM expression code space for $\mathbf{\psi}$ and learn a latent space $\mathds{R}^l$ for $\mathbf{w}$.

\vspace{2mm}
\noindent
\textbf{3DMM-Anchored Avatar Generative Model.} 
Inspired by Bai \etal \cite{monoavatar}, we adopt a 3DMM-anchored neural radiance field as our avatar representation (Fig. \ref{fig:network} right).
Specifically, instead of encoding all rendering information into a high-capacity neural network, local features are attached on the vertices of the 3DMM mesh scaffold reconstructed for the target identity and expression.
During rendering, each query point aggregates the features from k-Nearest-Neighbors (kNN) in the 3DMM vertices and sends it into a MLP network to predict color and density.
To simplify learning using existing 2D CNNs, the 3DMM vertex-attached features can be learned in the unified UV space and sampled using texture coordinates.

Our generative prior model generates identity and expression embeddings registered in the 3DMM UV space (Fig. \ref{fig:network} left).
In the identity branch, we use a StyleGAN2~\cite{stylegan2} generator to synthesize an identity feature map $\mathbf{F}_{\rm I}\in\mathds{R}^{h\times w \times c_i}$, where $h$ and $w$ are the resolution of the UV space, from a $l$ dimensional latent space.
In the expression branch, similar to MonoAvatar~\cite{monoavatar}, we train a convolutional-based U-Net to predict an expression feature map $\mathbf{F}_{\rm E}\in\mathds{R}^{h\times w \times c_e}$ from a vertex displacement map encoding per-vertex 3D deformation between 3DMM in neutral and the target expression.
The summation of two feature maps are then sampled by 3DMM vertices via texture coordinates to establish the 3DMM-anchored neural radiance field.
In practice, we set $h=w=256$, $c_i=c_e=64$, and $l=512$.

\vspace{2mm}
\noindent
\textbf{Auto-decoding Based Learning.}
We learn a generative avatar prior in an auto-decoding fashion~\cite{glo}.
Specifically, we assign each subject in our data a learnable latent vector, and jointly optimize them with the StyleGAN2-based~\cite{stylegan2} identity branch and the U-Net based expression branch.
Note that such training strategy can be highly unstable if very little data is available for each subject (\eg one image per subject like that in FFHQ~\cite{ffhq}) since the latent space $\mathbf{w}$ is highly under constrained.
With plenty of variation from various expression and camera viewpoints in our data, we directly apply the $\ell_1$ loss between the rendered images with the ground truth images without applying any latent space regularization loss. We observe that with a subject count exceeding approximately 2,000, the learned latent space exhibits smoothness and enables natural interpolation.

\subsection{Few-shot Adaptation}
Given the learned generative prior, we jointly adapt its weights and learn a new latent code $\mathbf{w}_{\rm target}$ assigned to a target identity given the $M$ views.

To make training stable and converge faster, we initialize the target latent code with the mean latent code across training subjects.
We follow the training schedule of PTI~\cite{pti} that alternatively performs model inversion to optimize $\mathbf{w}_{\rm target}$ and fine-tune the model weights $\theta$.

While the proposed generative prior model learning and adaptation enables few-shot avatar creation, we empirically find that this process can be unstable and is strongly dependent on the specific testing data. This is because the 3DMM fitting from a single view image is often not accurate in the presence of occlusions.

To alleviate this, we propose to jointly optimize camera pose and 3DMM facial expressions in addition to fine-tuning the model weights:
\begin{equation}
\mathbf{w}_{\rm target}, \theta_{\rm target} = \underset{\mathbf{w}, \theta, \{\Delta R_m, \Delta\mathbf{t}_m, \Delta\mathbf{\psi}_m\}}{\operatorname{argmin}} \mathcal{L}_{\text{recon}} + \mathcal{L}_{\text{reg}},
\end{equation}
where $\{\Delta R_m, \Delta\mathbf{t}_m, \Delta\mathbf{\psi}_m\}$ is additive updates to camera rotation, translation, and 3DMM expression code for each given image of the target identity, $m$ denotes image index,  $\mathcal{L}_{\text{recon}}$ is an $\ell_1$ loss between the rendered RGB values and the ground truth ones, and $\mathcal{L}_{\text{reg}}$ is a $\ell_2$ regularization loss on those updates:
\begin{equation}
\mathcal{L}_{\text{reg}} = \Sigma_m \|\Delta R_m \|_2^2 + \|\Delta \mathbf{t}_m \|_2^2 + \|\Delta \mathbf{\psi}_m \|_2^2
\end{equation}

\subsection{Training Schema}
Our prior model is trained on a multi-view dataset with photometric loss, where we penalize the $\ell_1$ distance between the rendering and the ground truth images. We optimize the prior model as an auto-decoder, where each identity has a latent code with 512 dimensions. For each training step, we randomly sample 256 pixels from 8 views of 4 expressions of 16 subjects, which yields a batch size of 131072. We train on 2407 subjects, each containing up to 13 different expressions and each expression contains 13 views. The prior model is trained on 16 GPUs for 1M steps, which takes around six days. We optimize our model using Adam~\cite{adam} with the learning rate set to 0.0005.

\begin{figure*}[t]
\centering
\includegraphics[width=\textwidth]{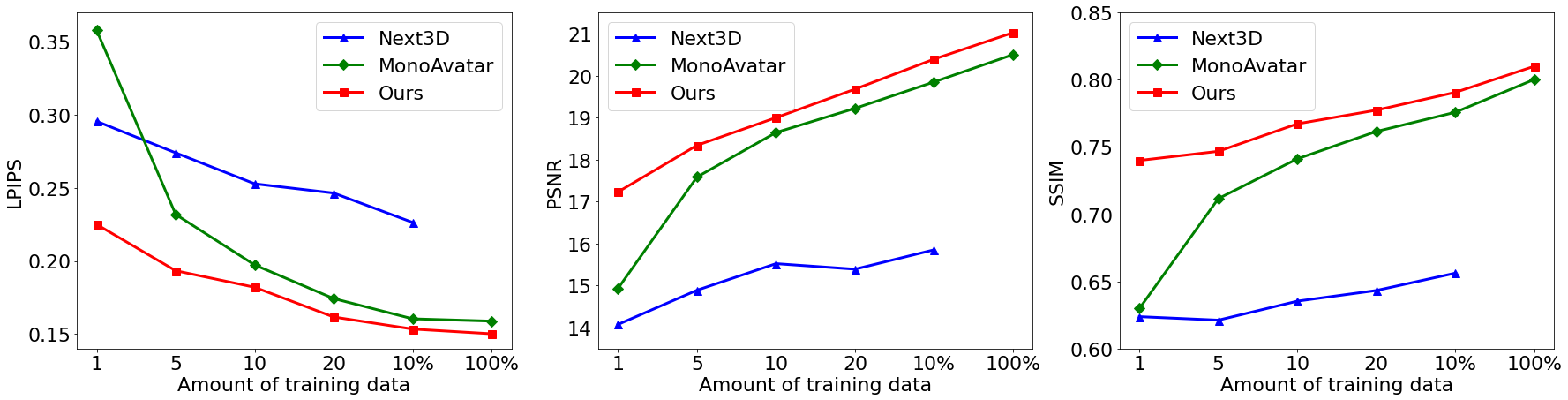}
\caption{New subject adaptation for different methods as varying \%  of training data. Our proposed method consistently outperforms the state-of-the-art approaches, particularly at the low data regime (e.g., one image). Even as the amount of training data increases, our method maintains its superior performance.}
\label{fig:curves}
\end{figure*}

\section{Experiments}
\label{sec:experiments}
In this section, we assess the performance of our approach under various scenarios, demonstrating the effectiveness of the learned latent representation to synthesize new subjects and its applicability for few-shot adaptation tasks.

\subsection{Evaluation Dataset, Metrics and Baselines}

\textbf{Monocular selfie-video dataset} We use a monocular RGB video dataset consisting of 6 subjects casually captured using smartphones and webcams, from MonoAvatar~\cite{monoavatar}. Each subject captures two distinct sessions for training and testing. In the training session, the subjects rotate their head as performing a series of expressions including extremes ones; in the testing session, the subjects performed expressions freely. Each subject contains $506-780$ frames of the training session and $320-603$ frames of the testing session, both being uniformly down-sampled from the video. Each frame consists of an RGB image with the background filtered out, a 3DMM fit and estimated camera parameters. For all the experiments, quantitative evaluations are performed on all testing frames for each subject. 

\noindent\textbf{Metrics} We employ the standard image quality metrics for our quantitative evaluations: Learned Perceptual Image Patch Similarity (LPIPS)~\cite{lpips}, Peak Signal-to-Noise Ratio (PSNR), and Structure Similarity Index (SSIM)~\cite{ssim}, following previous works~\cite{nerface,monoavatar}.

\label{sec:baseline}
\noindent\textbf{Baselines} In our experiments, we use the five baselines: 
(1) MonoAvatar~\cite{monoavatar} is a method to create a personalized 3D avatar using the same avatar backbone as ours.
(2) Next3D~\cite{next3d} is the state-of-the-art 3DMM conditional 3D-aware face generative model using tri-planes~\cite{eg3d} as avatar backbone. One-shot avatar creation has been demonstrated in their paper using PTI \cite{pti}, which can be extended to leverage more images when available.
(3) A variant of our method that uses the tri-planar representation (similar to Next3D) as a backbone with the avatar prior trained on our full dataset. We refer to this baseline as "Ours-TP".
(4) A variant of our model with the avatar prior trained on FFHQ~\cite{ffhq} dataset. We run an off-the-shelf 3DMM fitting algorithm~\cite{flame} to get the parametric model parameters for each individual image, and train our model treating each image as a unique identity. We refer to this baseline as "Ours-FFHQ". 
(5) A variant of our model with the avatar prior trained using single-view data from our captured dataset. Specifically, we randomly pick one out of the $13$ cameras for each subject, and use only the data from that camera for training. We refer to this baseline as "Ours-SV".

\subsection{Comparison on Few-shot Adaptation}
In this section, we compare our approach to existing few-shot avatar adaptation methods. We analyze the behavior of our model with respect to the available quantity of training data, number of training iterations and quality with respect to out of training expressions. Finally, we present an experiment showcasing the generative capabilities of our latent representation.


\begin{figure}[t]
\centering
\includegraphics[width=\linewidth]{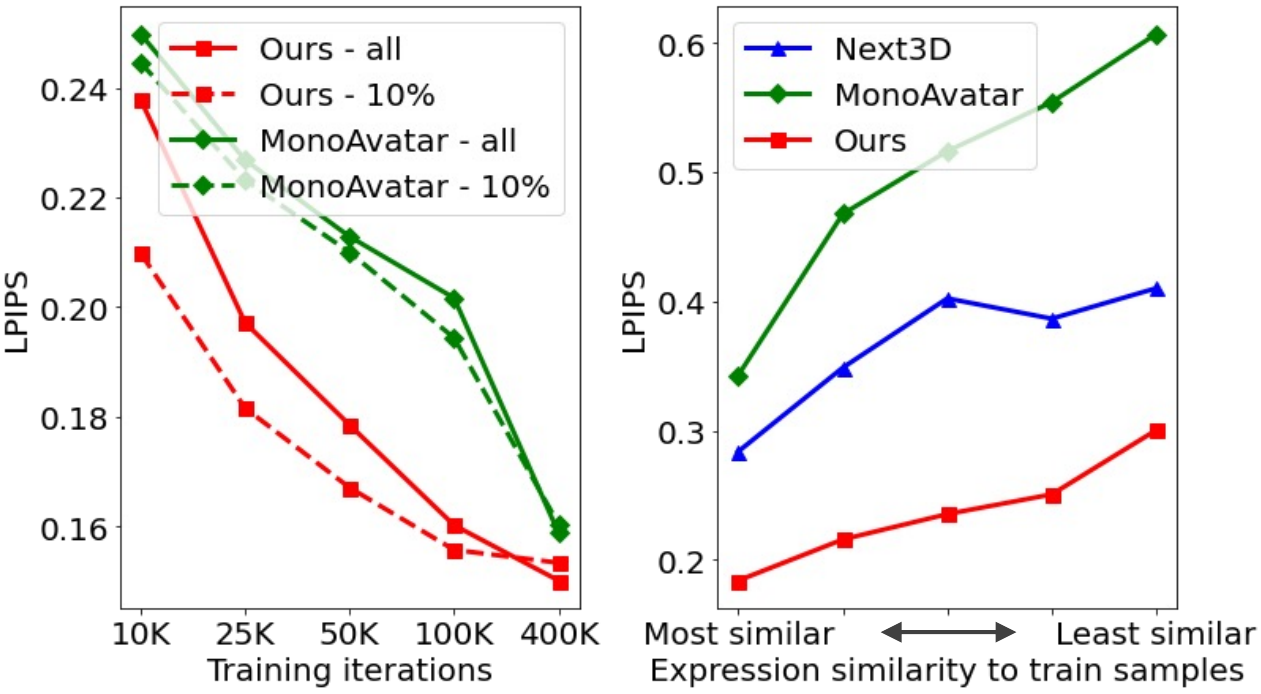}
\caption{Cross-method comparison against the training time (left) and the test images grouped in five bins: expressions of images in bins with larger index are less similar to the training image (right).}
\label{fig:curves_combined}
\end{figure}

\subsubsection{Few-shot Adaptation}
In this section, we evaluate the performance of our method on the few-shot avatar creation.
We compare with the state-of-the-art 3DMM conditional 3D-aware face generative model Next3D~\cite{next3d} and personalized avatar method MonoAvatar~\cite{monoavatar}.
For Next3D, we run PTI \cite{pti} with few-shot images on the official pre-trained model. MonoAvatar models are trained from scratch.

\vspace{2mm}
\noindent\textbf{Quality w.r.t quantity of training data.}
For each method, we create avatars for each subject using different number of images, \ie 1, 5, 10, 20, $10\%$, and $100\%$ of the training data.
For each case, we uniformly sample images from the training split and use the same set of images for all three methods.
We found the sampled images empirically cover a reasonable range of expression and camera viewpoints.

The quantitative comparison is shown in Fig.~\ref{fig:curves}.
Our method significantly outperforms the other methods especially at few-shot settings (\eg number of views is smaller than 10).
Both our method and MonoAvatar use more training images effectively with consistently better performance, but our method keeps a consistent performance gain over MonoAvatar even when using $100\%$ of the data.
This indicates that our generative prior also provides a good initialization of the network weights for learning the avatar.

Some qualitative results from one-shot and five-shot adaptation are shown in Fig.~\ref{fig:qual_comp}.
Our method produces avatar results with more accurate expression, more consistent identity, and less artifacts compared to Next3D and MonoAvatar.

\vspace{2mm}
\noindent\textbf{Quality w.r.t training time.}
For the settings that use 10\% or all training images where our result are relative close to MonoAvatar~\cite{monoavatar}, we additionally compare quality against training iterations and show results in Fig.~\ref{fig:curves_combined} (Left). It can be seen that our performance improves much faster than MonoAvatar at fewer iterations.
Our model achieves the same quality in fewer iterations (\eg our method achieves $LPIPS=0.18$ in $1/8$ the number of iterations compared to MonoAvatar), owing to the learnt latent space.

\vspace{2mm}
\noindent\textbf{Quality w.r.t testing expression.}
To illustrate the ability of our generative model to learn both identity and motion priors, we evaluate the performance of all three methods in the one-shot scenario on testing frames with varying the expression similarity (\ie $\ell_2$ distance between 3DMM expression codes) to the training frame. As depicted in Fig.~\ref{fig:curves_combined} (Right), all methods exhibit a decline in performance as the expression difference between the testing and training frames increases. However, our method demonstrates a more gradual decline compared to the other two approaches.


\begin{figure*}[t]
\centering
\includegraphics[width=\textwidth]{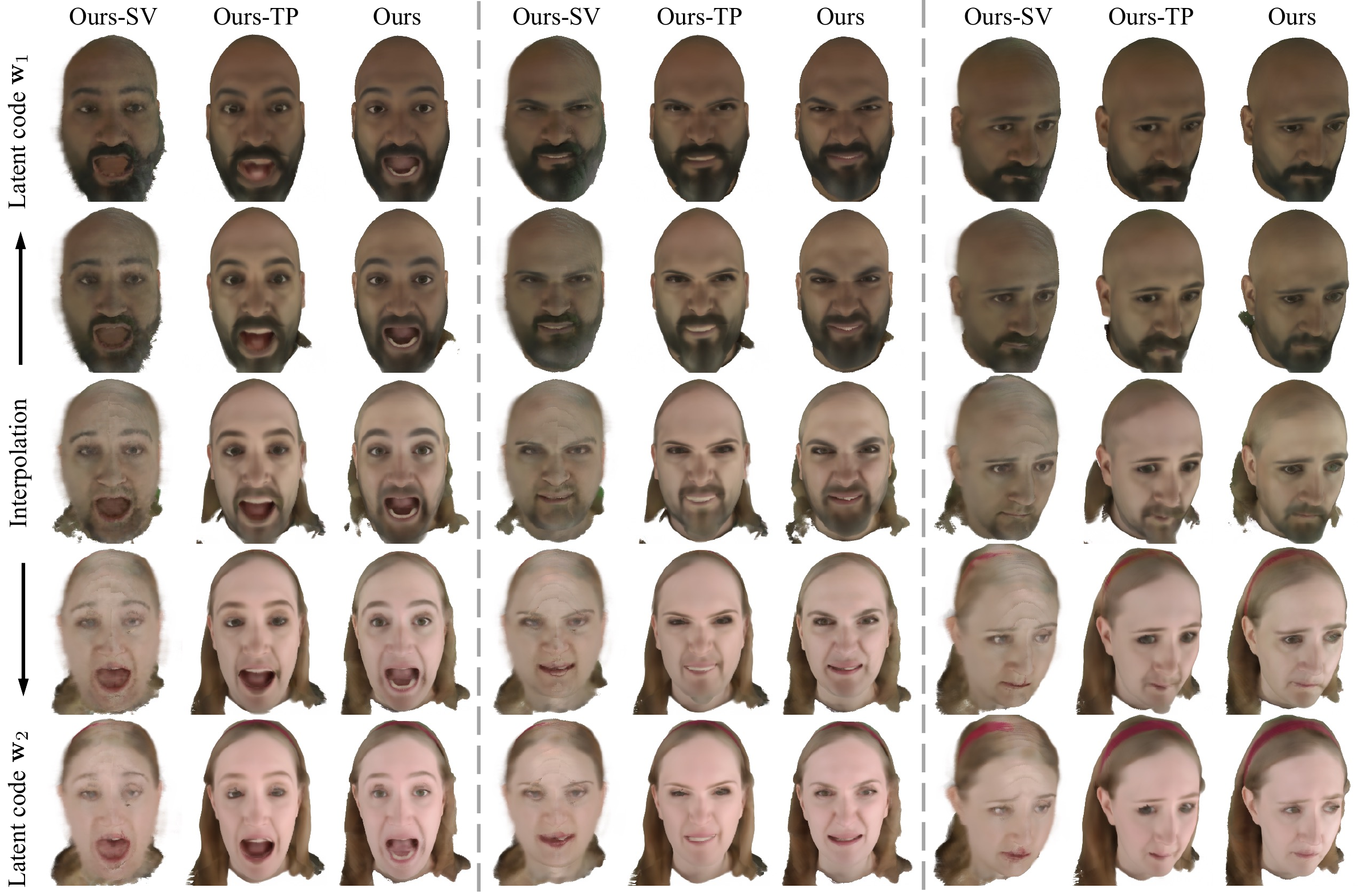}
\caption{Avatar generation in novel identity by interpolating latent code of two subjects. We compare the proposed approach with a Tri-planes Backbone (TP) and with a model trained on a single view (SV), showing how we can generate higher quality renderings of subjects and smoother interpolations.}
\label{fig:avatar_gen}
\end{figure*}

\subsubsection{Generative Avatar Prior}
Finally we briefly show the generative quality of our learned avatar prior. 
We linearly interpolate between two identity latent codes and show the avatar renderings in different expressions and camera viewpoints in Fig.~\ref{fig:avatar_gen}.
Our prior model learns a smooth and interpolatable latent space, and the generated avatar images show appealing visual quality and consistent identity across multiple controlling signals. Compared to the variant using tri-planes as avatar backbone (``Ours-TP") or single-view subset of our training data (``Ours-SV"), our full method generates more realistic and sharper results.

\begin{table}[h]
\centering
\scriptsize
\setlength\tabcolsep{5pt} 
\begin{tabular}{lcccc}
\toprule
& \textbf{One-shot} & \textbf{Five-shot}  \\
& LPIPS $\downarrow$ / PSNR $\uparrow$/ SSIM $\uparrow$  & LPIPS $\downarrow$ / PSNR $\uparrow$ / SSIM $\uparrow$   \\
\midrule
MonoAvatar~\cite{monoavatar} & 0.358 / 14.93 / 0.630 & 0.232 / 17.59 / 0.712 \\
\midrule
Next3D~\cite{next3d} & 0.296 / 14.08 / 0.624 & 0.274 / 14.89 / 0.621 \\
Ours-FFHQ & 0.234 / 17.16 / 0.737 & 0.207 / 18.07 / 0.745 \\
\midrule
Ours-SV & 0.251 / 16.97 / 0.730 & 0.208 / 18.15 / 0.745 \\
Ours-TP & 0.293 / 16.28 / 0.708 & 0.266 / 17.39 / 0.714 \\
Ours (w/o corr.) & 0.234 / 17.11 / 0.735 & 0.205 / 18.20 / 0.744 \\
Ours & \textbf{0.225} / \textbf{17.23} / \textbf{0.740} & \textbf{0.193} / \textbf{18.34} / \textbf{0.747} \\
\bottomrule
\end{tabular}
\caption{Ablation study. Note how our approach outperforms the other possible design choices by a considerable margin.}
\label{tab:ablation}
\end{table}

\subsection{Ablation Study}
In this section, we analyse the importance of data and network backbone in learning a good avatar prior. Additionally, we validate the efficacy of fitting correction in few-shot adaptation tasks. Results are presented in Tab.~\ref{tab:ablation}.

\begin{figure*}[th!]
\centering
    \begin{subfigure}[b]{\textwidth}
        \centering
        \includegraphics[width=\textwidth]{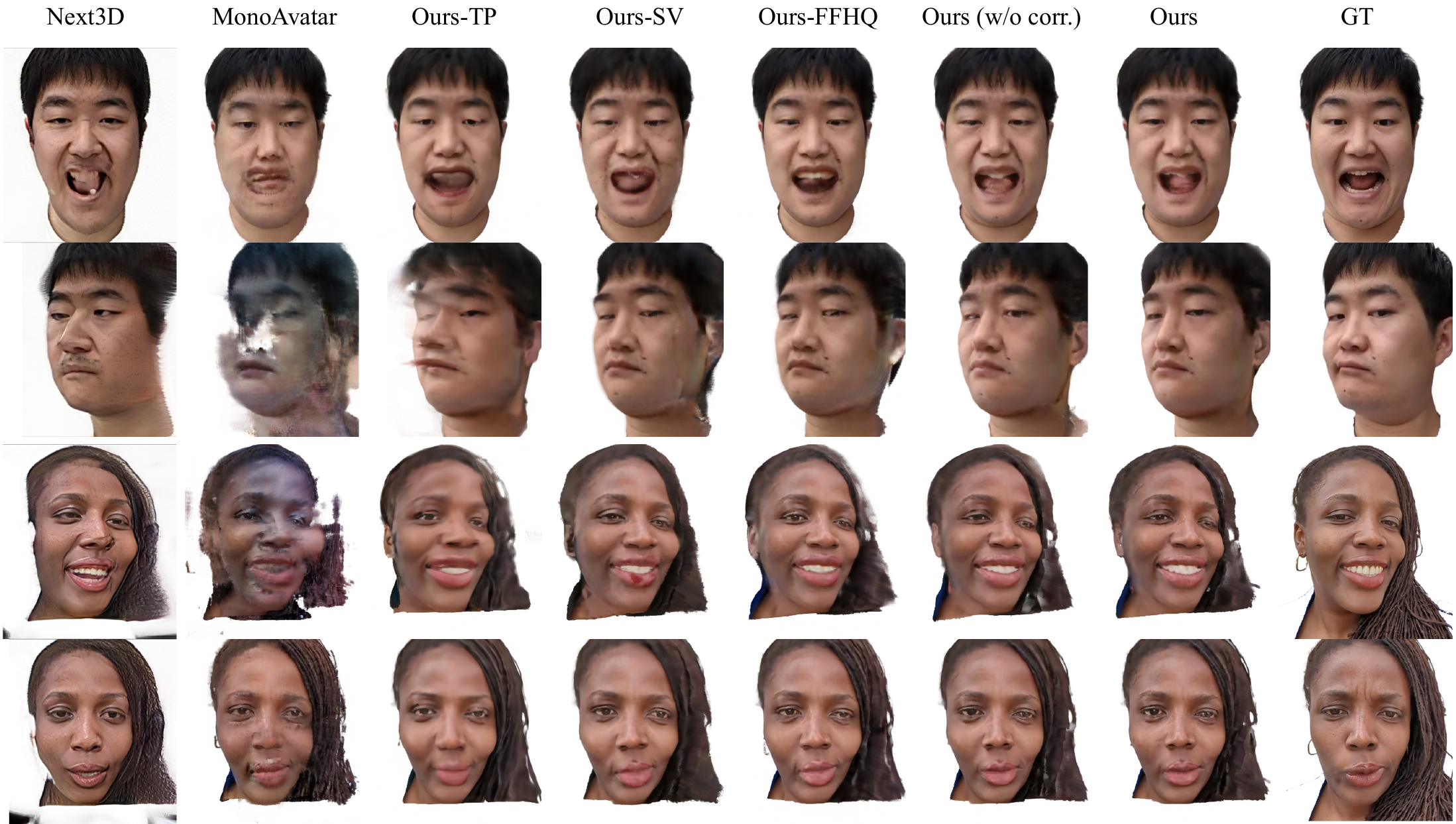}
        \vspace{-5mm}
        \caption*{(a) One-shot avatar creation}
    \end{subfigure}
    \par\medskip
    \begin{subfigure}[b]{\textwidth}
        \centering
        \includegraphics[width=\textwidth]{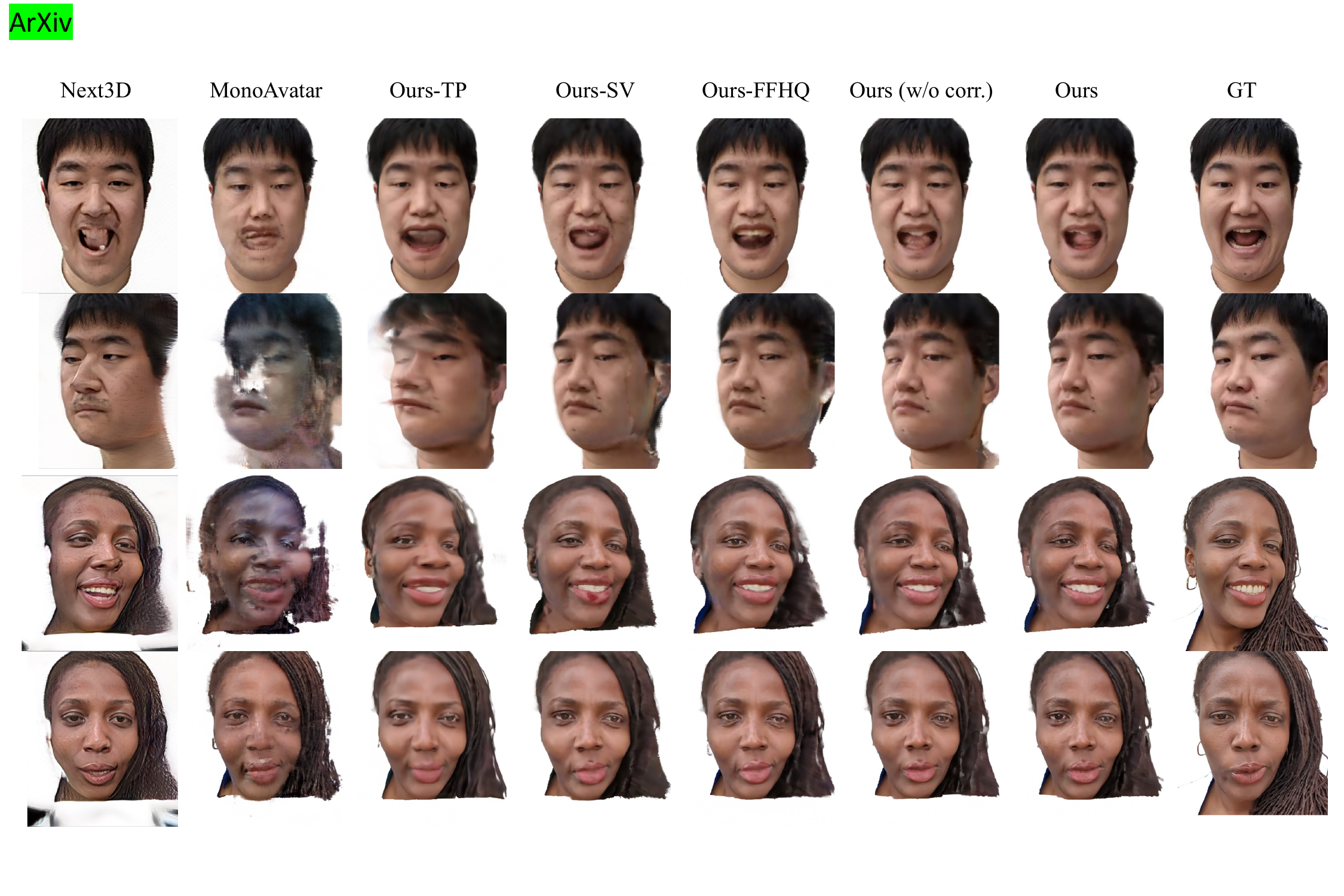}
        \caption*{(b) Five-shot avatar creation}
    \end{subfigure}
\caption{Qualitative comparisons of our approach against state of art methods such as Next3D \cite{next3d} and MonoAvatar \cite{monoavatar}. Additionally, we compare our approach to alternative design choices outlined in Sec.~\ref{sec:experiments}. As evident from the results, our model excels in capturing facial expressions and head poses with greater fidelity, both in one-shot (top) and five-shot (bottom) adaptation scenarios.}
\label{fig:qual_comp}
\end{figure*}

\paragraph{Does multi-view data help?}
We answer this question by performing few-shot adaptation on ``Ours-SV" and ``Ours-FFHQ" (refer to Sec.\ref{sec:baseline}), both of which are trained on single-view data, and compare with our model trained on multi-view data (``Ours"). The quantitative results are shown in Tab.~\ref{tab:ablation}. Our method outperforms them for all metrics under both one-shot and five-shot settings. Qualitative comparisons are shown in Fig.~\ref{fig:qual_comp}. Compared to ``Ours-SV" and ``Ours-FFHQ", ``Ours" trained with multiview data is more robust when head rotates and has less artifacts on the side face, \eg around the ear area.

\paragraph{Does avatar backbone help?}
We conduct this comparison both on our data and FFHQ dataset.
On our data, we perform few-shot adaptation using ``Ours-TP" (refer to Sec.\ref{sec:baseline}), which replaces the backbone of our model with the tri-planar representation \cite{next3d}.
On FFHQ, we directly compare ``Ours-FFHQ" with Next3D.
Quantitative results are shown in Tab.~\ref{tab:ablation}, and priors with our backbone consistently outperform tri-plane representation on both dataset and all metrics, showing the superiority of our backbone for avatar prior.

\paragraph{Does fitting correction help?}
We show a comparison in Tab.~\ref{tab:ablation} for few-shot adaptation with and without fitting correction. Jointly optimizing 3DMM fitting during adaptation helps achieve better overall performance.


\section{Conclusion}
\label{sec:conclusion}
We present a novel generative 3D implicit head avatar model using 3DMM-anchored radiance field representation that serves as a powerful prior for few-shot adaptation of new subjects.
We demonstrate that it is particularly critical to learn such a prior for dynamic avatar using multi-view and multi-expression data, instead of single-view data, in order to learn both animation and identity priors.
We also show that 3DMM-anchored neural radiance field is a more effective backbone compared to tri-planar representation for avatar creation through auto-decoding based on few-shot inputs.
To overcome unsatisfactory 3DMM fitting and camera calibration in few-shot adaptation, we showed that jointly optimizing the parametric face model fit with generative inverse fitting can significantly improve the performance.

{
    \small
    \bibliographystyle{ieeenat_fullname}
    \bibliography{main}
}

\clearpage
\setcounter{page}{1}
\maketitlesupplementary



\section{Multi-view Multi-Expression Face Capture Dataset}
\begin{figure}[h!]
\centering
\includegraphics[width=\linewidth]{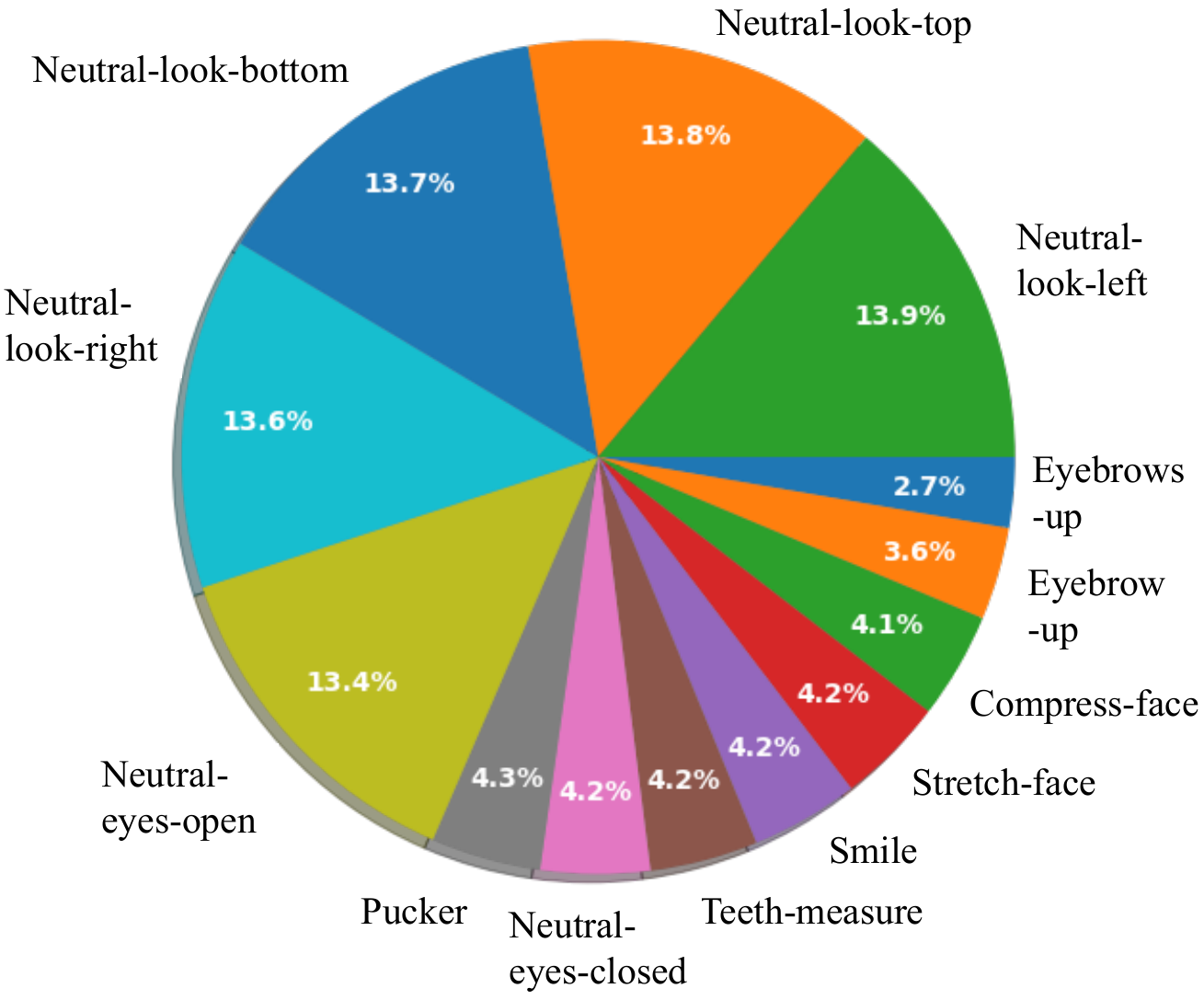}
\caption{Distribution of expressions in our multi-view multi-expression face capture dataset.}
\label{fig:dataset_dist}
\end{figure}

\begin{figure*}[t]
\centering
\includegraphics[width=\textwidth]{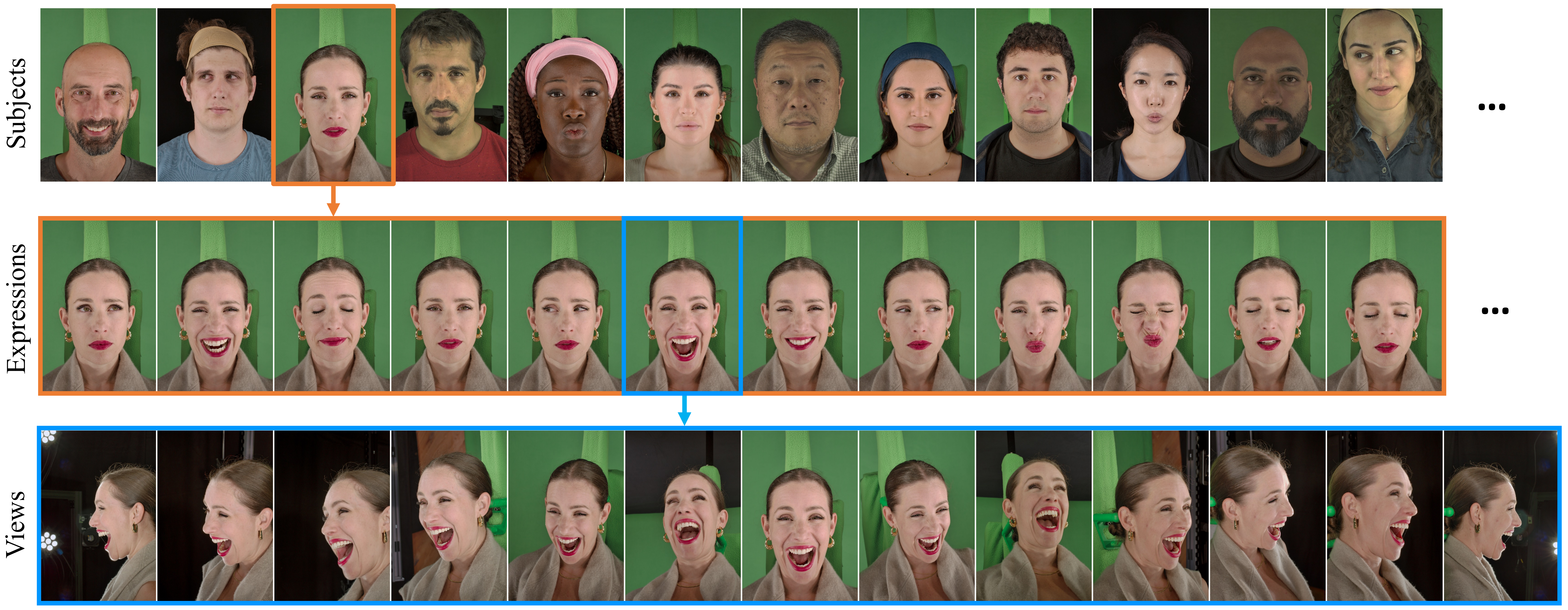}
\caption{We capture high-resolution facial images of a total $2407$ subjects in $13$ pre-defined facial expression from $13$ sparse camera viewpoints. The cameras are evenly spaced throughout the front hemisphere. 
}
\label{fig:dataset}
\end{figure*}

Our Multi-view Multi-Expression Face Capture Dataset is organized in a subject-expression-view hierarchy. We capture high-resolution facial images of a total 2407 subjects in 13 pre-defined facial expression from 13 sparse camera viewpoints. The cameras are evenly spaced throughout the front hemisphere. The statistics for the 13 pre-defined types of expressions are shown in Fig.~\ref{fig:dataset_dist}. We show examples of different subjects, expressions and views in Fig.~\ref{fig:dataset}.

\section{Identity Preservation in Few-shot Avatar Creation}
Preservation of target identity across different expressions and viewpoints is challenging for the task of avatar creation, especially when training data only contains one or few images. To explore the identity preservation performance of our model, we leverage a pre-trained face recognition network~\cite{arcface} to extract embeddings of rendered images and ground truth ones, and computing the cosine similarity between them as the metric for identity preservation~\cite{lan2023self}. The quantitative comparison with baselines methods under one-shot and five-shot settings are shown in Tab.~\ref{tab:idsim}. Ours achieves the best performance.

\begin{table}[h]
\centering
\small
\begin{tabular}{lcccc}
\toprule
& One-shot & Five-shot \\
\midrule
MonoAvatar~\cite{monoavatar} &0.513$\pm$0.155  &0.722$\pm$0.117  \\
\midrule
Next3D~\cite{next3d} &0.672$\pm$0.091  &0.695$\pm$0.095  \\
Ours-FFHQ &0.776$\pm$0.087  &0.790$\pm$0.093  \\
\midrule
Ours-SV &0.764$\pm$0.092  &0.787$\pm$0.092  \\
Ours-TP &0.685$\pm$0.114  &0.696$\pm$0.131  \\
Ours (w/o corr.) &0.779$\pm$0.085  &0.805$\pm$0.088  \\
Ours  &\textbf{0.782}$\pm$\textbf{0.083}  &\textbf{0.822}$\pm$\textbf{0.078}  \\
\bottomrule
\end{tabular}
\caption{Comparison with baseline methods on the metric of identity similarity. The numbers are the mean and standard deviation of the cosine similarity between the feature embeddings extracted from rendered images and ground truth ones.}
\label{tab:idsim}
\end{table}

\section{Effect of Number of Training Subjects on Generative Avatar Prior}
\begin{figure*}[t]
\centering
\includegraphics[width=\textwidth]{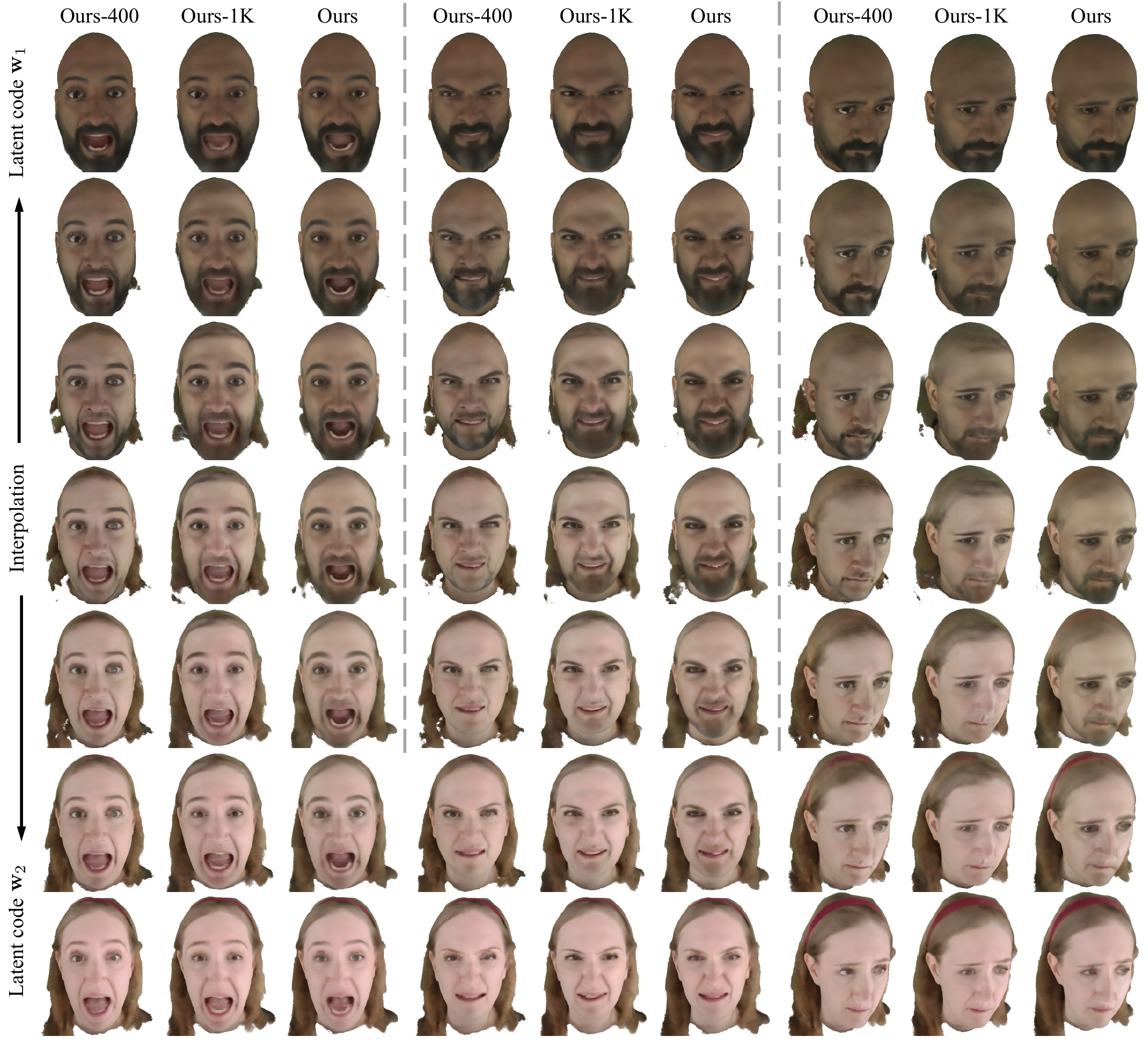}
\caption{Avatar generation in novel identity by interpolating latent code of two subjects. We compare our prior models trained with different number of subjects. "Ours-400" and "Ours-1K" denote prior models trained with 400 and 1000 subjects respectively. "Ours" is trained with all 2407 subjects. Note that ours produce a smoother transition between interpolations for skin color, beard and hair regions.}
\label{fig:avatar_gen_wrt_n_subs}
\end{figure*}

The generative quality of our learned avatar prior depends on the number of subjects in the training data. To understand this dependency, we train prior model on two subsets of our training data: 400 subjects and 1000 subjects. We then compare them with our prior trained using all 2407 subjects by linearly interpolating between two identity latent codes and show the avatar renderings in different expressions and camera viewpoints, as shown in Fig.~\ref{fig:avatar_gen_wrt_n_subs}. Note that the transitions between interpolations for prior models trained with fewer subjects are less smooth, especially for skin color, beard and hair regions.

\end{document}